\documentclass[final]{l4dc2026}

\title[Latent Control Barrier Functions]{%
How to Train Your Latent Control Barrier Function: \\ 
Smooth Safety Filtering Under Hard-to-Model Constraints}
\usepackage{times}

\author{%
 \Name{Kensuke Nakamura} \Email{kensuken@cmu.edu} \\
 \Name{Arun L. Bishop} \Email{arunleob@cmu.edu}\\
 \Name{Steven Man} \Email{stevenwman@cmu.edu}\\
  \Name{Aaron M. Johnson} \Email{amj1@cmu.edu}\\
   \Name{Zachary Manchester}  \Email{zacm@cmu.edu}\\
   \Name{Andrea Bajcsy} \Email{abajcsy@cmu.edu}\\
   \addr Carnegie Mellon University%
}
\usepackage{xcolor}
\usepackage{moreverb,url}
\usepackage{xspace}
\usepackage{amssymb}
\usepackage{amsmath}
\usepackage{amsbsy}
\usepackage{bbm}
\usepackage{dsfont}
\usepackage{soul}
\usepackage{wrapfig}
\usepackage{booktabs} %
\usepackage{algorithm}
\usepackage{algpseudocode}

\usepackage{subfiles}
\usepackage{csquotes}

\usepackage{array}

\usepackage{multirow}
\usepackage{graphicx}
\usepackage[table]{xcolor}
\usepackage{bm,upgreek}

\newcommand\BibTeX{{\rmfamily B\kern-.05em \textsc{i\kern-.025em b}\kern-.08em
T\kern-.1667em\lower.7ex\hbox{E}\kern-.125emX}}

\definecolor{wine}{RGB}{204, 0, 102}
\definecolor{magenta_wine}{RGB}{158, 44, 143}
\definecolor{dusty_wine}{RGB}{143, 59, 101}
\definecolor{ocean}{RGB}{13, 121, 202}
\definecolor{light_ocean}{RGB}{18, 178, 235}
\definecolor{dark_ocean}{RGB}{10, 89, 148}
\definecolor{grey}{RGB}{170, 170, 170}
\definecolor{light-grey}{RGB}{220, 220, 220}
\definecolor{dark_gray}{rgb}{0.2, 0.2, 0.2} 
\definecolor{med-grey}{rgb}{0.3, 0.3, 0.3} 
\definecolor{grape}{RGB}{112,48,160}
\definecolor{aqua}{RGB}{52,172,139}
\definecolor{dark_aqua}{RGB}{35,115,93}
\definecolor{dark_orange}{RGB}{216,92,0}
\definecolor{vibrant_orange}{RGB}{250, 160, 26}
\definecolor{vibrant_blue}{RGB}{14, 120, 255}
\definecolor{vibrant_pink}{RGB}{255, 0, 104}
\definecolor{dark_red}{RGB}{122, 0, 0}
\definecolor{dark_green}{RGB}{0, 92, 34}
\definecolor{dusty_blue}{RGB}{77, 91, 128}
\definecolor{dark_brown}{RGB}{125, 54, 36}

\newcommand{\para}[1]{\medskip\noindent\textbf{#1. }}

\newcommand{\draft}[1]{\normalsize{\color{magenta}#1}}

\newcommand{\new}[1]{\normalsize{\color{blue}#1}}

\newcommand{\ours}{\textcolor{black}{LatentCBF}\xspace}
\newcommand{\oursNonom}{\textcolor{black}{LatentCBF-NoMix}\xspace}

\newcommand{\oursNogp}{\textcolor{black}{LatentCBF-NoGP}\xspace}

\DeclareMathOperator*{\argmax}{argmax}
\DeclareMathOperator*{\argmin}{argmin}

\newcommand{\state}{s}
\newcommand{\statet}{\tilde{s}}
\newcommand{\stateSpace}{\mathcal{S}}
\newcommand{\dyns}{f}
\newcommand{\cbf}{B}

\usepackage{annotate-equations}
\newcommand{\buffer}{\mathcal{B}}

\newcommand{\latent}{z}
\newcommand{\latentSpace}{\mathcal{Z}}
\newcommand{\dynz}{f_{\latent}}
\newcommand{\dataset}{\mathcal{D}}

\newcommand{\ellparam}{\mu}

\newcommand{\obs}{o}
\newcommand{\obsSpace}{\mathcal{O}}
\newcommand{\enc}{\mathcal{E}}

\newcommand{\action}{a}
\newcommand{\actiont}{\tilde{a}}
\newcommand{\actionSpace}{\mathcal{A}}

\newcommand{\policy}{\pi}

\usepackage{fontawesome5}
\newcommand{\shield}{\text{\tiny{\faShield*}}}

\newcommand{\failure}{\mathcal{F}}

\newcommand{\marginfunc}{\ell}

\newcommand{\valfunc}{V^{\shield}}
\newcommand{\qfunc}{Q^{\shield}}

\newcommand{\gradientTarg}{\beta}

\newcommand{\fallback}{\policy^{\shield}} 
\newcommand{\piNom}{\policy^{\text{nom}}}

\newcommand{\datafail}{\mathcal{D}_\text{fail}}
\newcommand{\datanotfail}{\mathcal{D}_{\text{safe}}}

\begin{document}
\maketitle

\begin{abstract}%
Latent safety filters extend Hamilton-Jacobi (HJ) reachability to operate on latent state representations and dynamics learned directly from high-dimensional observations, enabling safe visuomotor control under hard-to-model constraints.
However, existing methods implement ``least-restrictive" filtering that discretely switch between nominal and safety policies, potentially undermining the task performance that makes modern visuomotor policies valuable. 
While reachability value functions can, in principle, be adapted to be control barrier functions (CBFs) for smooth optimization-based filtering, we theoretically and empirically show that current latent-space learning methods produce fundamentally incompatible value functions. 
We identify two sources of incompatibility: 
First, in HJ reachability, failures are encoded via a ``margin function'' in latent space, whose sign indicates whether or not a latent state is in the constraint set. However, representing the margin function as a classifier yields saturated value functions that exhibit discontinuous jumps. We prove that the value function's Lipschitz constant scales linearly with the margin function's Lipschitz constant, revealing that smooth CBFs require smooth margins. 
Second, reinforcement learning (RL) approximations trained solely on safety policy data yield inaccurate value estimates for nominal policy actions, precisely where CBF filtering needs them.
We propose the \textbf{LatentCBF}, which addresses both challenges through gradient penalties that lead to smooth margin functions without additional labeling, and a value-training procedure that mixes data from both nominal and safety policy distributions. Experiments on simulated benchmarks and hardware with a vision-based manipulation policy demonstrate that \textbf{LatentCBF} enables smooth safety filtering while doubling the task-completion rate over prior switching methods. Project Page: \url{https://cmu-intentlab.github.io/latent_cbf/}
\end{abstract}


\section{Introduction}
\begin{figure*}[t!]
    \centering
    \includegraphics[width=\textwidth]{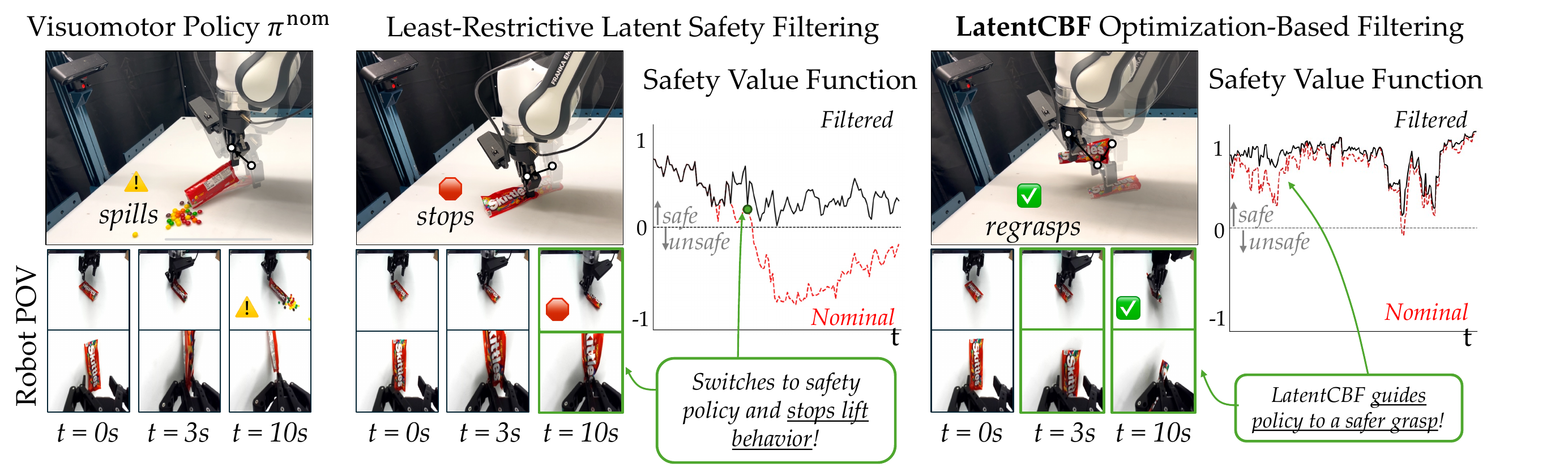}
    \vspace{-2em}
    \caption{\textbf{Safety Filtering a Visuomotor Manipulation Policy.} Both the nominal policy and the safety filters take as input the RGB images shown on the bottom. (left) Unfiltered nominal diffusion policy spills the bag's contents. (center) Least-restrictive latent safety filter prevents spilling, but also stops the diffusion policy from lifting the bag. (right) Our \textbf{LatentCBF} guides the diffusion policy to a safe grasp and it completes the pickup task.}
    \label{fig:front-fig}
    \vspace{-1.8em}
\end{figure*}

In theory, safety filters---such as Hamilton–Jacobi (HJ) reachability \citep{mitchell_time-dependent_2005}, control barrier functions (CBFs) \citep{ames_control_2017}, or model-predictive shielding \citep{bastani2021safe}---can monitor and correct \textit{any} nominal policy to prevent safety violations. Yet a substantial gap remains between theory and practice. One cause for this gap is a shift in how robot policies are designed. For example, visuomotor manipulation policies \citep{chi2024diffusionpolicy, intelligence2025pi} increasingly operate end-to-end, performing complex tasks directly from RGB camera inputs. Such policies are deployed in conditions that violate many assumptions underlying classical safety filters: state representations are complex and partially observable (e.g., deformables), dynamics models or simulators may be unavailable, and the safety constraints are often extremely hard to specify analytically (e.g., spilling, as seen in left of Figure~\ref{fig:front-fig}).

To align a safety filter's assumptions with those of a nominal visuomotor policy, recent work introduced \textit{latent safety filters} that operate on learned latent state spaces and dynamics (world) models trained directly from high-dimensional observations \citep{nakamura_generalizing_2025}.
These methods represent safety constraints as the level set of a ``margin function,'' implemented as a classifier trained on labeled failure observations whose sign indicates a failure state.
An approximate HJ-reachability problem is then solved in the latent space to derive a safety value function and fallback policy, enabling visuomotor filters that, for example, prevent a manipulator from spilling a bag’s contents using only wrist and third-person RGB images (Figure~\ref{fig:front-fig}). However, existing approaches rely on ``least-restrictive" filtering that discretely switches between nominal and safety policies, degrading the nominal policy’s performance. 
A natural alternative is optimization-based filtering, as in CBF methods \citep{ames2019control}, to minimally alter the nominal policy while maintaining safety. 
Since recent work has adapted HJ-based value functions into CBF-like formulations \citep{choi2021robust, tonkens2022refining, oh2025safety}, it is compelling to try to use the latent-space value function directly for such filtering.

\begin{figure}[t!]
    \centering
    {\includegraphics[width=0.99\textwidth]{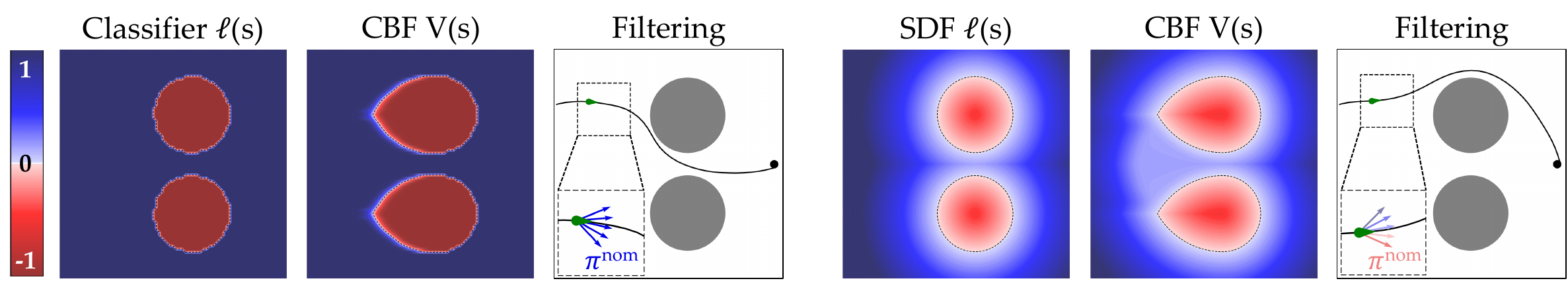}}
    \vspace{-0.5cm}
    \caption{\textbf{CBFs as a Function of the Margin Function.} %
    Even with a perfect model, a classifier-based $\marginfunc(\state)$ yields a CBF with poor signal during action filtering (left). A smooth %
    margin function %
    provides a rich signal for the CBF to evaluate alternative actions (right). %
    }
    \label{fig:dubins_priv}
    \vspace{-2em}
\end{figure}

In this paper, we theoretically and empirically show why existing latent safety filters are \textit{incompatible} with smooth, CBF-style filtering in latent space, and we propose algorithmic solutions that make such filtering feasible in practice. 
The incompatibility arises from how reachability-based value functions are approximated in latent space. First, encoding constraints via a classifier prevents the HJ value function from learning smooth gradients, leaving safety filters unable to assess relative action safety until the last moment (see Figure \ref{fig:dubins_priv}). 
Second, in high-dimensional latent spaces ($\geq512$), actor–critic reinforcement learning (RL) is typically used to approximate the safety value function. 
However, this creates a distribution mismatch at deployment: the value function, trained on conservative data from the safety policy, is used to evaluate a nominal policy that explores different states and actions, yielding poor value estimates where CBF-style filters need them the most.

Our contributions address both sources of incompatibility and yield latent discrete-time CBFs (called \textbf{\ours}) suitable for optimization-based safety filtering directly from observations (Figure~\ref{fig:front-fig}, right). First, we prove that, under mild conditions, the Lipschitz constant of the safety value function scales linearly with that of the margin function, showing that smooth value functions require smooth margin functions---a property violated by binary classifiers. This insight motivates our second contribution: a Wasserstein GAN-inspired \citep{arjovsky2017wasserstein, gulrajani2017improved} training method that learns smooth margin functions “for free,” using only the same binary safe/fail labels as done in prior work. Third, we introduce a mixed exploration strategy that blends state–action samples from both nominal and safety policies, ensuring accurate value estimates across the regions critical for CBF filtering. We evaluate \ours in a benchmark simulation with privileged CBF access and in a vision-based Franka manipulation task, finding that it enables $45\%$ smoother interventions in simulation and doubles ($38\% \rightarrow 80\%$) the safe-task success rate of visuomotor manipulation policies on hardware compared to least-restrictive safety filters.

\section{Mathematical Background}
\label{sec:background}

Consider the discrete-time system governed by bounded dynamics $\state_{t+1} = \dyns(\state_t, \action_t)$ where $\state_t \in \stateSpace$ is the state of the system at time $t$ and the action $\action_t \in \actionSpace$ is selected from a bounded control set, $\actionSpace$. 
Let the set of states that are already in failure (i.e., the system has already violated safety) be denoted by the failure set: $\failure \subset \stateSpace$. For example, $\failure$ could represent states where a robot vehicle is in collision, or a state where the contents of a bag have already spilled out during robot manipulation.

\begin{definition}[Discrete-time Control Barrier Function \citep{agrawal2017discrete}]

    \noindent A function $\cbf: \stateSpace \rightarrow \mathbb{R}$ is a discrete-time control barrier function (CBF) if $\forall \state \in \stateSpace$ the function $\cbf$ satisfies\footnote{In full generality, the right side of the inequality in \eqref{eq:dcbf_cond} can be an extended class $\kappa_\infty$ function. For ease of presentation, we restrict the notation to linear functions of $\cbf(\state)$ parameterized by $\alpha$.} 
\begin{equation}
    \exists \action \in \actionSpace \text{ s.t. } 
    \cbf(f(s,a)) \geq \alpha \cbf(\state) 
    , \text{ for } \alpha \in [0, 1),   
    \label{eq:dcbf_cond}
\end{equation}
and $\Omega \cap \failure = \emptyset$ where $\Omega := \{ \state ~|~ \cbf(\state) > 0 \}$.%
\end{definition}
The condition \eqref{eq:dcbf_cond} renders the zero superlevel set $\Omega$ a \textit{control invariant safe set}. This means for any state $\state \in \Omega$, then there exists an action $\action$ s.t. $\dyns(\state, \action) \in \Omega$. Since this control invariant set  does not intersect with $\failure$, states in this set can avoid entering $\failure$ for all future time. 
Discrete-time CBF filters find the minimal action adjustment satisfying these constraints through the following optimization:
\begin{equation}
\label{eq:dcbf_filtering}
    \action^\star = \argmin_{\action \in \actionSpace} || \action - \piNom(\state)||, ~~~~ \text{s.t. } 
    \cbf(f(\state, \action)) \geq \alpha 
    \cbf(\state)
\end{equation}
where $\alpha \in [0, 1)$ is a parameter that dictates how ``quickly'' the safety filter will begin to override the nominal policy $\piNom$ as the system approaches the boundary of the control invariant set.

\para{Hamilton-Jacobi (HJ) Reachability} 
While the optimization problem in \eqref{eq:dcbf_filtering} defines a ``minimally invasive'' safety filter problem, a well-known drawback of CBFs is the challenge of constructing a valid function $\cbf(\cdot)$ that satisfies Definition 1.  %
We build upon a recent line of work \citep{choi2021robust, tonkens2022refining, oh2025safety} that proposes the use of Hamilton-Jacobi value functions as a CBF\footnote{These works relax the discrete-time CBF constraint to hold only for $\{ \state \in \stateSpace ~|~ \cbf(\state) \geq 0 \}$, which still ensures control invariance of the safe set at the expense of the set attractivity property of traditional CBFs \citep{cortez2021robust}.}. %
HJ reachability begins by defining a \textit{margin function} $\marginfunc(\state): \stateSpace \rightarrow \mathbb{R}$, which implicitly defines the failure set\footnote{In general, the zero-superlevel set of $\marginfunc(\state)$ is \textit{not} control-invariant, and thus cannot be used directly as a CBF.} as the zero-sublevel set $\failure := \{ \state ~|~ \marginfunc(\state) < 0\}$. For example, in hand-designed state spaces, this is often a signed distance function to the boundary of the constraint. 
Then one can obtain a safety value function $\valfunc: \stateSpace \rightarrow \mathbb{R}$ satisfying the discrete-time Hamilton-Jacobi fixed-point equation \citep{mitchell_time-dependent_2005, fisac_bridging_2019}:
    \begin{equation}
    \label{eq:hj}
        \valfunc(\state) = \min \{ \marginfunc(\state), \max_{\action \in \actionSpace} \valfunc(\dyns(\state, \action)) \},
    \end{equation}
with a zero-superlevel set representing the \textit{maximal} control-invariant safe set. Mathematically, $\valfunc(\state) \geq 0 \iff \state \in \Omega$  and $\state \notin \failure$, and $\valfunc$ satisfies %
\eqref{eq:dcbf_cond}
for all states in its zero-superlevel set \citep{oh2025safety}. This implies that $\valfunc(\state)$ computed via reachability-based techniques can be used as a discrete-time CBF (i.e., $\cbf(\state) = \valfunc(\state)$).
However, solving \eqref{eq:hj} faces the \textit{curse of dimensionality}, as computation grows exponentially with state dimension.
To address this, neural approximations \citep{bansal_deepreach_2020, hsu_isaacs_2024} have emerged, which, despite lacking formal guarantees, scale safety filtering to high-dimensional systems with strong empirical safety.

\section{Problem Formulation: Safety Filtering via Latent Control Barrier Functions}
\label{sec:problem}
Unlike traditional safety filters from 
Section~\ref{sec:background} that assume access to fully-observed dynamical systems, we do \textit{not} assume access to a hand-designed model of $\state \in \stateSpace$, dynamics $\dyns(\state, \action)$, or the failure set $\failure$. 
Instead, we only assume access to a dataset of trajectories consisting of observations, $\obs_t \in \obsSpace$, %
(e.g., RGB images and proprioceptive state), and robot actions, $\action_t \in \actionSpace$%
. Let this dataset of diverse observation-action trajectories be $\mathcal{D} = \{(\obs_t, \action_t, \obs_{t+1})_i\}_{i=1}^N$ collected from the robot interacting with its environment. 
This setting is common modern visuomotor policies such as diffusion or flow-matching policies \citep{chi2024diffusionpolicy, intelligence2025pi}. %
We also assume that we can identify if $\state_t \in \failure$ solely from the observation $\obs_t$
(e.g., whether the contents of a bag spilled on the table) and have access to binary labels indicating as such.
Our goal is to bring optimization-based filtering, like that of CBFs, closer to the vision-based capabilities of modern visuomotor policies. 
To do this we leverage the paradigm of latent world models.

\para{Latent State \& Dynamics} Latent world models \citep{ha2018worldmodels} jointly learn a latent state space representation $\latent \in \latentSpace$, an encoder from observations to latent states $\enc(\obs): \obsSpace \rightarrow \latentSpace$, and a (potentially stochastic) dynamics model $\dynz: \latentSpace \times \actionSpace \rightarrow \Delta(\latentSpace)$, where $\Delta$ denotes a distribution, given real-world robot dataset, $\mathcal{D}$. %
For brevity, we only introduce the necessary mathematical models here and provide additional background in the Appendix.

\para{Latent-Space Safe Control}
Prior work has shown it is possible to approximately solve \eqref{eq:hj} in latent spaces using actor-critic reinforcement learning \citep{nakamura_generalizing_2025}. %
These methods approximate the safety state-action value function $\qfunc:\latentSpace \times \actionSpace \rightarrow \mathbb{R}$ (critic) and a learned safety fallback policy $\fallback : \latentSpace \rightarrow \actionSpace$ (actor) via the latent time-discounted HJ fixed point equation:
\begin{equation}
\label{eq:latent_hj}
        \qfunc(\latent, \action) = (1-\gamma)\marginfunc(\latent) + \gamma \min \Big\{ \marginfunc(\latent), ~ \max_{\action' \in \actionSpace} \mathbb{E}_{\latent' \sim \dynz(\latent, \action)}\qfunc(\latent', \action') \Big\},
\end{equation}
where the safety fallback policy $\fallback(\latent)$ learns $\argmax_{\action \in \actionSpace} \qfunc(\latent, \action)$. Here, $\gamma \in [0,1)$ is the discount factor that recovers \eqref{eq:hj} as $\gamma \rightarrow 1$, %
in the sense that $\valfunc(\latent) = \max_{\action \in \actionSpace} \qfunc(\latent, \action)$ \citep{fisac_bridging_2019}.%

\para{Latent Discrete-Time CBF} At runtime, our goal is to solve the following latent discrete-time CBF optimization problem to minimally steer any visuomotor policy $\piNom: \obsSpace \rightarrow \actionSpace$ %
such that it can effectively perform its task while staying within the zero-superlevel set of $\valfunc(\latent)$: 
\begin{equation}
\label{eq:latent_cbf}
    \action^\star = \argmin_{\action \in \actionSpace} || \action - \piNom(\obs)||, ~~~~ \text{s.t. } 
    \qfunc(\latent, \action) - \epsilon \geq \alpha \left( \qfunc(\latent,\fallback(\latent)) - \epsilon\right), 
\end{equation}
where $\latent = \enc(\obs)$ is obtained by querying the world model's encoder on the same observations that are passed into the nominal visuomotor policy. The hyperparameter $\epsilon$ is a small positive constant used to account for any learning inaccuracies of the zero-level set. Note that, if we query the state-action value function with the \textit{safety fallback policy's} action, then we obtain $\valfunc(\latent) \approx \qfunc(\latent,\fallback(\latent))$; this is the ``safest'' we could ever be at this latent state. 
If we want to evaluate the safety of any \textit{other} action $\action \in \actionSpace$, we can query  $\qfunc(\latent,\action) \approx \valfunc(\dynz(\latent,\action))$, which forms the left-hand side of the inequality constraint. 
Thus, we reformulate the CBF-style safety filter from \eqref{eq:dcbf_filtering} where the latent state-action value function learned via \eqref{eq:latent_hj} is our discrete-time CBF \citep{oh2025safety}. %

\section{The Theory-Practice Gap for Latent Control Barrier Functions}
\label{sec:theory-practice-gap}

Recall that, for HJ reachability to compute the discrete-time CBF via \eqref{eq:latent_hj}, we need %
a margin function $\marginfunc(\latent)$ that encodes the set of latent states that ``appear'' to be in failure, i.e., $\marginfunc(\latent) \leq 0 \iff \latent \in \failure_\latent$.
How can we obtain the latent margin function from a dataset of observation-action tuples in $\mathcal{D}$? 
Prior work \citep{nakamura_generalizing_2025} asks a stakeholder to look at the observations $\obs \in \mathcal{D}$ and assign binary labels to those that look like they are in failure (e.g., contents of the bag are spilled on the table). 
This is because, in practice, it is far easier to label an image from the robot's point of view (as shown in the bottom row of Figure~\ref{fig:front-fig}) as a failure or not than to provide per-frame and real-valued labels indicating \textit{how close} the robot is to failure. 

\para{Classification-Based Latent Margin Functions} This labeling procedure results in a dataset of $\latent^+ \in \datanotfail$ and $\latent^- \in \datafail$ for safe and failed latent states, respectively, obtained from labeled observations encoded via the world model's encoder $\enc(\obs) = \latent$. This is used to train a classifier that learns to discriminate latent states coming from visually failed observations from visually safe ones. The training objective for a bounded\footnote{To ensure the existence and uniqueness of the solution of \eqref{eq:latent_hj}, the margin function $\marginfunc{_\ellparam}(\latent)$ must be bounded, which can be enforced by using $\tanh(\cdot)$ as the final activation of the margin function or clipping its final output.}  classifier-based  $\marginfunc_\ellparam(\latent)$ with trainable parameters $\mu$ is: 
\begin{equation}
    \mathcal{L}^\delta_\text{sign}(\mu) = \mathbb{E}_{\latent^+ \sim \datanotfail}\big[\min\{0, \delta - \marginfunc_\ellparam(\latent^+)\}\big] + \mathbb{E}_{\latent^- \sim \datafail}\big[ \min\{0,\delta + \marginfunc_\ellparam(\latent^-)\}\big], 
     \label{eq:failure-classifier-loss}
\end{equation}
where parameter $\delta \in \mathbb{R}^+$ prevents $\marginfunc_\mu(\latent)$ from converging to degenerate solution. 
Since we assume that we can identify if $\state \in \failure$ from the observation $\obs$, this encourages $\marginfunc_\ellparam(\latent) \leq 0 \iff \state \in \failure$. 

\para{Challenge 1: Smooth Value Functions Need Smooth Margin Functions} In theory, the loss function in \eqref{eq:failure-classifier-loss} is sufficient for obtaining a margin function that makes the HJ fixed point solution from \eqref{eq:hj} well-defined \citep{bellman1952theory}. In practice, however, the margin function has a large Lipschitz constant due to the near-discrete jumps at the boundary of the failure set. 
Together with the need for $\marginfunc_\mu(\latent)$ to be bounded, this results in ``saturated'' value functions uninformative for solving \eqref{eq:latent_cbf}. 
We show  %
a classifier-based margin function %
and the resulting discrete-time CBF obtained via grid-based numerical solvers \citep{stanfordasl_hj_reachability} for a 3D Dubins' car (details in \ref{sec:sim}). 
During optimization-based filtering, saturated gradients prevent the CBF from assessing how action changes affect long-term safety (center, Figure~\ref{fig:dubins_priv}, where all sampled actions yield similarly high safety values).
In contrast, a smooth margin function produces a more sensitive CBF that distinguishes safer from riskier actions, scoring roughly half of the sampled actions as less safe (right, Figure~\ref{fig:dubins_priv}).

We characterize this relationship by proving that even under the perfect state and deterministic dynamics, the Lipschitz constant $L_{\valfunc}$ of the discounted HJ value function scales \textit{linearly} in the Lipschitz constant $L_\marginfunc$ of the margin function.

\begin{definition}[Lipschitz Continuity]
    A function $f: \mathcal{S} \to \mathbb{R}$ is \emph{Lipschitz continuous} if there exists a constant $L_f \in \mathbb{R}_{\ge 0}$ %
    such that
        $|f(s) - f(\tilde{s})| \leq L_f \, \|s - \tilde{s}\|, \quad \forall s, \tilde{s} \in \mathcal{S}.$
    The smallest such $L_f$ is called the \emph{Lipschitz constant} of $f$.
\end{definition}

\begin{theorem}[Margin-to-Value Lipschitz Bound]
    Let the margin function $\marginfunc(\state)$ and time discounted HJ value function $\valfunc(\state)$ be Lipschitz continuous with constants $L_\marginfunc$ and $L_{\valfunc}$,  respectively. Let the discrete-time dynamics $\dyns(\state, \action)$ be uniformly Lipschitz in $\state$ with constant $L_\dyns$ such that for a fixed discount factor $\gamma \in [0,1)$, $\gamma L_\dyns < 1$.
    Then the Lipschitz constant of $\valfunc(\state)$ scales linearly in $L_\marginfunc$ :%
    \[ 
    L_{\valfunc} \leq L_\marginfunc \cdot \max \left\{1, \frac{1-\gamma}{1 - \gamma L_\dyns} \right\}. 
    \]
\end{theorem}

\noindent \textbf{Proof:} See Appendix~\ref{subsec:app-proof-bound}.

\para{Challenge 2: Distribution Mismatch Between Safety and Nominal Policies} 
Even with a smooth margin function, another practical issue arises when reinforcement learning (RL) is used to approximate the safety value function via \eqref{eq:latent_hj}.
Prior works use actor-critic RL algorithms which jointly learn a safety policy (actor) $\fallback_\nu(\latent)$ with parameters $\nu$ and critic (value function) $\qfunc_\phi(\latent, \action)$ with parameters $\phi$ by iteratively ``rolling out'' the current actor policy within a world model and fitting the critic \citep{Sutton1998}. 
Specifically, at each step within the world model, a transition $(\latent, \action, l, \latent')$ is saved in a replay buffer, $\buffer$, that stores a dataset of past latents, actions, and the margin function label $l = \marginfunc_\mu(\latent)$.  
The critic is trained via supervised learning to the Bellman equation target:
\begin{equation}
\small
\label{eq:bellman_target}
    \mathcal{L}_{\text{critic}}(\phi) = \mathbb{E}_{(\latent, \action, l, \latent') \sim \buffer} \big[(\qfunc_\phi(\latent, \action) - y_{\text{target}})^2 \big], ~~ y_{\text{target}} = (1-\gamma) l + \gamma \min\{l, \qfunc_\phi(\latent', \action') \},
\end{equation}
where $\action' = \fallback_\nu(\latent')$. 
Given the current estimated critic, the actor is then optimized by minimizing the loss $\mathcal{L}_{\text{actor}}(\nu) = \mathbb{E}_{\latent \sim \buffer} \big[-\qfunc_\phi(\latent, \fallback_\nu(\latent))\big]$.
Here is where we can see our second challenge. 
Since the replay buffer $\buffer$ contains \textit{only safe actions and latent transitions} obtained from $\fallback$ (e.g., actions that may never grasp the bag), the critic only learns a high-quality estimate of how far the safety policy can get from the failure set. However, the critic rarely evaluates the safety outcomes of any other task-oriented action (e.g., dragging the bag along the table), since these state-action samples are rarely visited and put into the buffer $\buffer$. 
This off-the-shelf training process is at odds with how the critic will be used at deployment in CBF-style filtering, where we must evaluate the safety of task-relevant actions near $\piNom$---actions the critic has likely \textit{never} encountered during training.

\section{How to Train Your Latent Control Barrier Function}
\label{sec:how-to-train}

Motivated by our analysis in Section~\ref{sec:theory-practice-gap}, we propose two algorithmic modifications to make latent HJ value functions informative latent CBFs. First, we improve the optimization landscape of \eqref{eq:latent_cbf} by reducing the Lipschitz constant of the margin function $\marginfunc_{\ellparam}(\latent)$, and use knowledge of a nominal policy $\piNom$ to diversify the state-action coverage during critic learning to improve the value estimates. 

\para{Smooth Margin Functions via Smooth Safety Discriminators}
Our key idea for learning smooth margins \textit{without dense supervision} is to draw inspiration from Wasserstein GANs (WGAN) \citep{arjovsky2017wasserstein}.
This method learns a smooth discriminator (i.e., our margin function $\marginfunc_\ellparam(\latent)$) that distinguishes between two classes of samples by regularizing its Lipschitz constant. 
We employ the objective function from 
WGAN-GP~\citep{gulrajani2017improved}:
\tikzset{annotate equations/text/.style={font=\fontfamily{ppl}\small}}
\begin{equation}
\label{eq:wgan}
\small
    \mathcal{L}_\text{WGAN}(\mu)
    =
     \lambda_{\text{w}} \cdot \Big( \eqnmarkbox[dark_green]{w}{\mathbb{E}_{\latent^- \sim \datafail}[\marginfunc_\mu(\latent^-)]
    - \mathbb{E}_{\latent^+ \sim \datanotfail}[\marginfunc_\mu(\latent^+)]} \Big) + \lambda_{\text{gp}} \cdot  \mathbb{E}_{\hat{\latent} \sim \mathcal{D}_\text{interp}}
    \big[\eqnmarkbox[red]{gp}{(\|\nabla_{\hat{\latent}} \marginfunc_\mu(\hat{\latent})\|_2 - \gradientTarg)^2}\big],
\end{equation}
\annotate[yshift=-0.35em]{below,left}{w}{$\marginfunc{_\ellparam}(\latent)$ ranks safe $>$ unsafe}
\annotate[yshift=-0.35em]{below,left}{gp}{Penalize norm of the gradient}

\noindent where $\hat{\latent} \sim \mathcal{D}_\text{interp}$ is obtained by sampling $\latent^+ \sim \datanotfail$, and $\latent^- \sim \datafail$ and linearly interpolating $\hat{\latent}=\eta\latent^-+(1-\eta)\latent^+$ such that $\eta \sim U(0,1)$. This objective encourages $\marginfunc_\mu(\latent)$ to assign higher values to safe samples while the gradient penalty objective regularizes the Lipschitz constant toward $\beta \in \mathbb{R}_{>0}$ over straight lines connecting safe and unsafe latent states \citep{gulrajani2017improved}.

Note, however, that the margin function which minimizes \eqref{eq:wgan} is defined only up to an additive constant and will \textit{not} result in a zero sublevel set which semantically corresponds to $\failure$. %
We remedy this by including the sign loss $\mathcal{L}_\text{sign}^\delta$ from \eqref{eq:failure-classifier-loss} with $\delta = 0$ since the WGAN loss already prevents a degenerate $\marginfunc_\mu(\latent)$.
This results in the overall loss function for training our latent margin function:
\begin{equation}
\vspace{-0.5em}
\label{eq:gp_margin}
    \mathcal{L} = \mathcal{L}_\text{WGAN} + \lambda_{\text{sign}} \cdot \mathcal{L}^{\delta=0}_\text{sign}.
\end{equation}
where $\lambda_{\text{w}}, \lambda_{\text{gp}}, \text{and } \lambda_{\text{sign}}$ are positive scalars that balance the contribution of the loss terms\new{\footnote{This function is trained in isolation without a separate generator network, bypassing unstable  adversarial training.}}.

\para{Mixing Safety \& Nominal Policy Trajectories to Address Distribution Mismatch}
We propose a simple modification to the actor-critical RL pipeline to address the distributional issues that impair the quality of $\qfunc(\latent, \action)$ when queried with actions that differ from $\fallback(\latent)$. 
We populate the replay buffer $\buffer$ with \textit{an equal proportion} of safety-oriented trajectories induced by the co-optimized actor $\fallback$ and trajectories generated by a task-oriented policy\footnote{We assume that the task-oriented policy is given to us and does not change over the course of RL training.} $\piNom$. 
At each timestep, we store transitions of the form $(\latent, \action, l, \latent', \action') \in \buffer$, where both $\action$ and $\action'$ are sampled from the same policy (either $\fallback$ or $\piNom$) to fit the critic\footnote{Unlike approaches that learn a CBF via policy evaluation \citep{so2024train}, our approach still performs full deep RL.}. This enables the critic to learn safety estimates that remain accurate for task-relevant actions likely to be encountered during deployment.

\para{Sampling-based Safety Filtering via LatentCBFs}
Unlike the continuous-time CBF with control-affine dynamics, the discrete-time CBF optimization problem in \eqref{eq:latent_cbf} does \textit{not} admit a quadratic program and is in general nonconvex, making it challenging to optimize efficiently in high-dimensional action spaces. To make \eqref{eq:latent_cbf} tractable we use zeroth-order optimization and sample the space of actions $\actionSpace$ using a mixture of $\piNom$ and $\fallback$, from which we create a subset that satisfy the CBF constraint, $\actionSpace_{\text{CBF-Safe}}$. The filter returns the most similar task-driven action from $\actionSpace_{\text{CBF-Safe}}$. 
Evaluating the objective and constraints of 
\eqref{eq:latent_cbf} %
can be accelerated with the parallelization capabilities of modern hardware (e.g., the entire process takes $10$ ms for 7.6k samples for our 7DOF manipulator). If no valid sample exists, we default to $\fallback$. Additional details can be found in Appendix \ref{sec:sampling_appendix}.

\section{Simulation \& Hardware Experiments}
We conduct simulation and hardware experiments to evaluate each component of \textbf{\ours} for safety filtering. 
All experiments assume access to a labeled offline dataset $\dataset$, a latent world model (e.g., $\enc(\obs), \dynz(\latent, \action)$) trained on this data, and a nominal diffusion policy $\piNom(\obs)$ that performs task-oriented actions from raw images. $\piNom$ is trained with both safe and unsafe demonstrations to test how effectively our safety filters guide an erroneous visuomotor policy. Details in Appendix  \ref{subsec:implementation-details}.

\subsection{Simulation: Vision-Based Dubins' Car Navigation}
\label{sec:sim}

Let a Dubins' car system have state $\state = (x, y, \theta)$ given by its position and steering angle $\theta$ and continuous-time dynamics $\dot{\state} = [\cos(\theta), \sin(\theta), \action]$ where 
 $\action \in [-2,2]$ is turn rate. The dynamics are discretized using RK4 and timestep $dt = 0.1$. The $x$ and $y$ positions are bounded in $[-1.5, 1.5]$. The true failure set $\failure$ is two circles centered at $(0.25, 0.65)$ and $(0.25, -0.65)$, with radius $r = 0.5$. Observations $\obs \in \obsSpace$ are RGB images and the angle $\theta$ (see Figure \ref{fig:dubins_priv}). 
 We label ground truth failures using privileged simulator information. 
 We train two margin functions and respective value functions using our proposed method (\textbf{GP}) and a baseline $(\textbf{NoGP})$ from \citep{nakamura_generalizing_2025}.

\para{Result: Gradient Penalties Yield Smoother Margin Functions Without Extra Labels}
We report the F1 score and smoothness of each margin function evaluated over 100 trajectories generated by $\piNom$ without filtering. Smoothness is measured as the maximum single-timestep change in $\ell(\latent)$ within a trajectory, i.e.,   $\max_t|\marginfunc{_\ellparam}(\latent_{t+1}) - \marginfunc{_\ellparam}(\latent_t)|$. Although both methods use the same dataset, \textbf{GP} reduces the largest margin function gradients by $86\%$ from $1.2 \pm 0.76$ to $0.17 \pm 0.065$ while maintaining a similar F1 score to \textbf{NoGP} (\textbf{GP}: $0.991$ vs. \textbf{NoGP}: $0.981$).

\begin{table*}[h!]
\centering
\setlength{\tabcolsep}{6pt}
\renewcommand{\arraystretch}{0.7}
\begin{tabular}{l|cc|cc|c}
\toprule
\multirow{3}{*}{\textbf{Safety Filter}} 
& \multicolumn{2}{c|}{\textbf{NoGP}} 
& \multicolumn{2}{c|}{\textbf{GP}} 
& \textbf{None} \\
\cmidrule(lr){2-3} \cmidrule(lr){4-5} \cmidrule(lr){6-6}
& Avg. ${|\Delta\action|}$ & Safety Rate 
& Avg. ${|\Delta\action|}$ & Safety Rate 
& Safety Rate \\
\midrule
LR   & $1.5\pm1.1$ & $97\%$ & $2.0\pm1.3$ & $100\%$ & \multirow{2}{*}{$41\%$} \\
CBF  & $1.3\pm1.0$ & $97\%$ & $1.1\pm0.9$ & $100\%$ &  \\
\bottomrule
\end{tabular}
\caption{\textbf{Simulation: On the Quality of Latent Safety Filters.}
CBFs trained with gradient-penalties yield smoother interventions, decreasing the average override magnitude relative to LR by $45\%$ compared to only $13\%$ for the CBF without gradient penalties.}
\label{tab:cbf_dubins_wide}
\vspace{-0.8cm}
\end{table*}

\para{Result: LatentCBFs Perturb the Nominal Policy Less While Staying Safe}
We compare two CBF filters against least-restrictive (LR) filtering: $\action^\star =  \mathds{1}_{\{\qfunc(\latent, \piNom)< \epsilon\}} \cdot \fallback$ $+  \mathds{1}_{\{\qfunc(\latent, \piNom) \geq \epsilon\}} \cdot \piNom$ for both \textbf{GP} and \textbf{NoGP} value functions with $\epsilon=0.2$.
Whenever the filter modifies the action, we record the action difference with respect to the nominal policy $|\Delta \action| = |\action^\star - \piNom|$, referred to below as the action override magnitude. We also measure the safety rate via $\%$ of trajectories without a collision between the car and obstacles.  
For both methods, we ablate the effect of the gradient penalty on $\marginfunc{_\ellparam}(\latent)$ and report the results across 100 trajectories in Table \ref{tab:cbf_dubins_wide}. The nominal policy only achieves a $41\%$ safety rate. We find that CBF-style filtering decreases the average override magnitude no matter how the margin function is shaped; however, the CBF trained with \textbf{GP} yields a $45\%$ decrease in action override magnitude relative to the corresponding LR filter. The \textbf{NoGP} CBF %
only reduces action magnitude by $13 \%$ relative to the LR filter.
Even with the reduced action override magnitude, the CBF methods maintain high safety rates matching their LR counterpart.%

\subsection{Hardware: Safety Filtering a Visuomotor Manipulation Policy}
We next scale our method to a manipulation task from \cite{nakamura_generalizing_2025} where a robot arm must pick up an opened bag of small round candies (Skittles) without spilling them. 
The robot has a 7-D action space $\action \in \actionSpace$ consisting of %
end-effector (EEF) translation and axis-angle rotation, along with binary gripper actions. 
The observation space $\obs \in \obsSpace$ consists of the EEF pose and RGB images from a wrist-mounted and 3rd person camera (see %
Figure~\ref{fig:front-fig}). We use DINO-WM, a vision transformer that predicts DINOv3 embeddings \citep{zhou_dino-wm_2025} as our world model.

\para{Result: \ours Needs Data from $\piNom$ and $\fallback$ to Effectively Filter High-D Actions}%
We train two value functions: using our RL strategy from Section~\ref{sec:how-to-train}
(\textbf{\ours}) and a baseline 
that does not learn from task-oriented trajectories (\textbf{\oursNonom}), both with gradient penalties. %
The robot is initialized with its end-effector grasping the closed end of a bag resting on the table and executes an \textit{unsafe} open-loop diagonal lift. All CBF filters are deployed with $\alpha \in \{0.85, 0.925\}$ and compared against the unfiltered sequence (\textbf{No Filter}) and a least-restrictive safety filter (\textbf{Least Restrictive}) using the same value function as \textbf{\ours} with $\epsilon=0.05$.
\begin{figure}
    \centering    \includegraphics[width=1.0\linewidth]{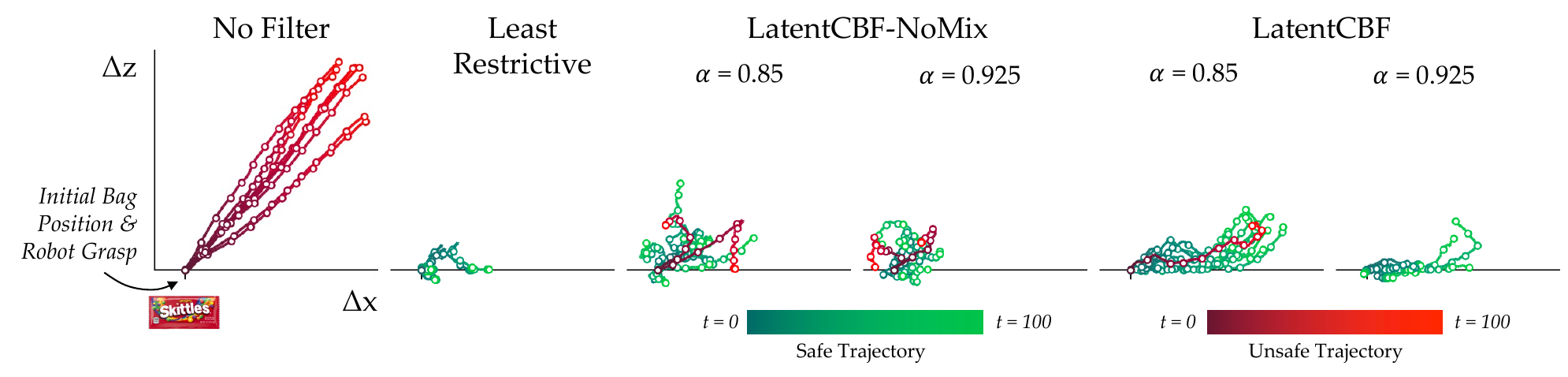}
    \vspace{-1cm}
    \caption{\textbf{Hardware: End-effector Trajectories in the X-Z Plane.} %
    \ours maintains control authority and safety compared to LR and unfiltered baselines. Without fitting the critic as in Sec. \ref{sec:how-to-train}, \oursNonom %
    permits erratic actions and spills in 20$\%$ of trials.}
    \vspace{-1cm}
    \label{fig:qualitative_hw}
\end{figure}
Figure \ref{fig:qualitative_hw} shows 10 end-effector trajectories in the X–Z plane for each method. \textbf{No Filter} always lifts and spills the bag. The \textbf{Least Restrictive} filter offers limited control: it lifts the bag the least ($+\Delta z$) and allows minimal translation along the table ($+\Delta x$). In contrast, \textbf{\ours} projects away unsafe action components and slides the bag along the table. Without mixing trajectories from $\piNom$ and $\fallback$ during RL training, \draft{\textbf{\oursNonom}} produces erratic actions and more safety violations than \textbf{\ours}.

\para{Result: LatentCBF Removes Unsafe Visuomotor Policy Modes Without Hindering the Task}
Finally, we evaluate our method when deployed in-the-loop with a visuomotor nominal policy. 
Here, $\piNom(\action \mid \obs)$ is the previously trained diffusion policy, which has both safe and unsafe interaction modes. 
We compare filtering this policy with three methods: the proposed \textbf{\ours}, a baseline \textbf{\oursNogp} (trained 
\textit{without} the gradient penalty for the margin function), and the \textbf{Least Restrictive} filter (the switching filter from the previous subsection).
For each method, we run 20 trials from three initial end-effector and bag configurations designed to elicit different nominal policy behaviors: IC1-consistently \textit{unsafe} actions, IC2-consistently \textit{safe} actions, and IC3-\textit{multimodal} grasping behavior.
We measure the percentage of spills (Fail), safe trials where the robot does \textit{not} lift the bag (Stall), and safely lifting without a spill (Success).

Table \ref{tab:cbf_dp} shows that none of the filters are overly conservative, preserving $\piNom$'s success rate in IC2. In IC1 and IC3, \textbf{\ours} reduces failures while improving task success, whereas baseline methods reduce failures at the cost of increased stalling. In IC3, \textbf{\ours} occasionally drives the policy out-of-distribution (OOD) of the imitation-based visuomotor policy, which lowers task success despite preventing failures. Overall, \textbf{\ours} more effectively guides the diffusion policy toward safe and successful lifts (80\% vs. 38\%).

\begin{table}[t!]
\centering
\small
\renewcommand{\arraystretch}{0.9}
\setlength{\tabcolsep}{3pt}
\begin{tabular}{c|c|ccc|ccc|ccc|ccc}
\toprule
\multirow{2}{*}{\textbf{Safety Filter}} & \multirow{2}{*}{\textbf{$\marginfunc{_\ellparam}(\latent)$}} & \multicolumn{3}{c|}{\textbf{Init Cond 1} (\%)} & \multicolumn{3}{c|}{\textbf{Init Cond 2} (\%)} & \multicolumn{3}{c|}{\textbf{Init Cond 3} (\%)} & \multicolumn{3}{c}{\textbf{Aggregate} (\%)} \\
 &  & Fail & Stall & Success & Fail & Stall & Success & Fail & Stall & Success & Fail & Stall & Success \\
\midrule
None   & -- & 100 & 0 & 0 & 0 & 0 & 100 & 85 & 0 & 15 & 62 & 0 & 38 \\
\arrayrulecolor{black!30}\midrule\arrayrulecolor{black}
LR  & GP & 0 & 90 & 10 & 0 & 0 & 100 & 0 & 95 & 5 & 0 & 62 & 38 \\
\arrayrulecolor{black!30}\midrule\arrayrulecolor{black}
\multirow{2}{*}{CBF} & No GP & 0 & 100 & 0 & 0 & 0 & 100 & 0 & 85 & 15 & 0 & 62 & 38 \\
  & GP & 0 & 0 & 100 & 0 & 0 & 100 & 0 & 60 & 40 & 0 & 20 & 80 \\
\bottomrule
\end{tabular}

\caption{\textbf{Hardware: Safely Guiding a Diffusion Policy.}  
Results are averaged over three initial conditions with 20 trials each. All safety filters prevent spilling, but \textbf{LR} and \textbf{\oursNogp} do so at the cost of task performance, whereas \textbf{\ours} more consistently steers the policy toward both safe and successful executions.}
\label{tab:cbf_dp}
\vspace{-1.5em}
\end{table}

\para{Result: Model-Free Filtering Scales to Thousands of Action Samples} Since the action search space grows exponentially with dimension, our method must scale to many samples for high-dimensional manipulation tasks. We consider two filtering schemes: (1) \textit{model-based}, which predicts $\latent' = \dynz(\latent, \action)$ and evaluates $\qfunc(\latent', \fallback(\latent'))$, and (2) \textit{model-free} (ours),
which directly evaluates $\qfunc(\latent, \action)$.
In Table \ref{tab:inference} we show that even with parallelization, model-based filtering is bottlenecked by inference through the large vision-transformer-based DINO-WM (with 19M parameters), preventing evaluation of $50+$ action samples on a NVIDIA A6000 ADA GPU with 48GB of VRAM. In contrast, model-free filtering scales to thousands of samples and takes about $10$ ms. 
Exact details of our sampling and filtering procedure are provided in Appendix \ref{subsec:sampling}.

\begin{table}[t!]
\centering
\renewcommand{\arraystretch}{0.9}
\setlength{\tabcolsep}{3pt}
\begin{tabular}{l|ccccc}
\toprule
& \multicolumn{5}{c}{\textbf{Filtering Speed (ms) as a Function of \# Samples}} \\
\# Samples & 10 & 30 & 50 & 3,800 & 7,600  \\
\midrule
Model-based &  40.70 $\pm$ 0.76 & 114.6 $\pm$ 2.6  & \texttt{out-of-mem} & \texttt{out-of-mem}  & \texttt{out-of-mem}  \\
Model-free &  0.33 $\pm$ 0.12 & 0.32 $\pm$ 0.08 & 1.85 $\pm$ 0.32 & 3.96 $\pm$ 0.19 & 9.60 $\pm$ 0.57  \\
\bottomrule
\end{tabular}
\vspace{-0.5em}
\caption{\textbf{Hardware: LatentCBF Filtering Speed.} Model-based filtering is infeasible for over $50$ actions (out of memory). Model-free can evaluate thousands of actions in $10$ ms. 
}
\label{tab:inference}
\vspace{-1.0cm}
\end{table}

\section{Conclusion \& Limitations}
In this paper, we identify two key limitations in existing latent safety filters: non-smooth value functions from classifier-based margins and inaccurate estimates due to training–deployment distribution mismatch. Our \textbf{LatentCBF} addresses both through Lipschitz-constrained margin learning inspired by Wasserstein GANs and mixed-policy value-function training. \textbf{LatentCBF} enables optimization-based safety filtering from high-dimensional observations, and experiments show that both contributions are critical for achieving safe yet minimally invasive filtering.

\para{Limitations} 
While \textbf{\ours} scales up safety filtering for visuomotor policies, it is not without limitations. Its reliance on learned latent representations, dynamics, and RL approximations precludes formal safety guarantees, which is an future important direction \citep{lutkus2025latent}. Successful filtering also depends on the base policy; if the CBF pushes the system into OOD states, the nominal policy may be unable to complete the task \citep{romer2026demonstrationssafedeploymentpathconsistent}. Future work should integrate uncertainty estimation or OOD detection \citep{seo2025uncertainty, sun2025latentpolicybarrierlearning}.

\bibliography{references.bib, l4dc2026}

\newpage

\section{Acknowledgments}
KN would like to acknowledge Oswin So for discussions on policy neural CBFs and inspiring the title of this work. He also thanks Gokul Swamy for helpful discussions surrounding reinforcement learning theory.

\section{Appendix}

\subsection{Extended Related Work}
\label{subsec:app-related-work}

\para{Scaling Safety Filters: From High-D States to Observations} Motivated by the difficulty of synthesizing control barrier functions by hand and numerically computing solutions to HJ reachability problems, an increasingly popular body of work involves scaling up safety filter synthesis with neural approximations. A plethora of methods for this exist, see \citep{wabersich_data-driven_2023} for a comprehensive survey. Neural CBF methods \citep{peruffo2021automated, zhao2020synthesizing, srinivasan2020synthesis} directly optimize the CBF constraint over a dataset of safe and unsafe samples which typically suffer from handling input constraints in practice \citep{dawson2023safe}. Alternative methods compute safe control-invariant sets with input constraints by approximating solutions to HJ reachability problems via self-supervised learning \citep{bansal_deepreach_2020, kim2025reachabilitybarriernetworkslearning} or reinforcement learning \citep{fisac_bridging_2019, hsu_isaacs_2024}. Similar to our work, is \citep{so2024train} which learn \textit{policy neural control barrier functions}; value functions obtained via policy evaluation with respect to a Hamilton-Jacobi Bellman equation. This method explicitly constructs a CBF around a nominal policy, but parametrizes the CBF as a purely state-value function and thus requires a forward simulation of the dynamics to evaluate the CBF constraint. Our work instead parameterizes the CBF as a state-action value function as in \citep{oh2025safety} which eliminates the need to forward simulation but adds the complexity of learning accurate state-action safety estimates. In general, all aforementioned works require access to explicit knowledge of the system state, dynamics, and failure constraints; an assumption relaxed in this work. Prior work synthesizing latent control barrier functions follow traditional neural CBF approaches \citep{kumar2024latentcbf, anand2025safetycertificationlatentspace}, requiring labels corresponding to knowledge of a \textit{safe control invariant set} which is assumed to be available but in practice difficult to find \citep{so2024train}. In contrast to our work which only requires labels corresponding to the \textit{failure set} and automatically synthesizes the CBF %
through reachability analysis.

\noindent \textit{Observations:} Our work is not the first to incorporate observation as an input to safety filters: prior work have used LIDAR as an input to the safety value function \citep{lin2024one, he2024agile} which can be trained in simulation and effectively deployed to real. Prior work has also explored simulated RGB data and conducted sim2real for least-restrictive safety filters that can operate directly from RGB \citep{hsu2023sim, chen2021safe}. While our work trains a latent world model for forward simulating the environment,  \citep{tong2023enforcing} trains a NERF representation raw perception data for forward simulating the control-barrier function condition. However these representations are more poorly suited for constraints beyond collision avoidance where the robot interacts with the environment via manipulation.

\para{Safe Latent-space Control} While our work primarily focuses on latent safety filtering following \cite{nakamura_generalizing_2025}, safe latent-space control does not necessarily have to be implemented as a filter. Prior work has explored using latent-space constraints to learn a task-oriented policy that is safe \textit{during} RL training \cite{zhan2024statewise, he2021androidsdreamelectricfences} or model-based planning in the latent-space \cite{wilcox22a} instead of learning a policy via RL. In contrast, this work assumes a nominal visuomotor policy is given a priori and is fixed. Our contribution in this work is the synthesis of a latent control barrier function that can safeguard this base policy even when operating directly from high-dimensional observations.

\para{Critic Learning in Reinforcement Learning} The primary concern in reinforcement learning is obtaining a performant \textit{policy}. Thus, the quality of the critic in actor-critic methods typically only matters insofar as it provides an effective source of supervision for optimizing this policy. While out-of-distribution actions are known to produce poor value estimates (the problem of overestimation bias), existing approaches to this problem instead introduce \textit{underestimation} \citep{fujimoto2018addressing} or \textit{conservative} critic updates \citep{kumar2020conservative} that maintain high-performance without addressing the inaccuracies of the critic itself. This is a non-issue in standard sum-of-reward RL that cares only about a highly performant policy but is unfavorable in our setting where we wish to use the critic for safety filtering. We draw inspiration from Hybrid RL \citep{song2023hybrid} which enables learning a critic from an offline dataset consisting of trajectories of varying quality, and online interaction. In our setting, we fit our critic using a mixture of both the safety fallback actions from $\fallback$ and nominal task-oriented actions from $\piNom$.

\subsection{Implementation Details}
\label{subsec:implementation-details}

\para{Simulation: World Model and Nominal Policy} For our simulated Dubins' car experiment, we use an open source implementation of the recurrent state space model \citep{hafner2025mastering} in PyTorch \cite{dreamerv3-torch} and a diffusion policy following the implementation from \cite{chi2024diffusionpolicy}. We list all hyperparameters in Table \ref{tab:dubins_hyperparams} and \ref{tab:diffusion}. These models take as input a (128x128x3) RGB image of the environment, and the steering angle $\theta$ of the system. For learning the margin function, we parameterize it as an MLP with two hidden layers [512, 512] and SiLU activations. We apply the $\texttt{stopgrad}[\latent]$ to the latent states when optimizing the margin function with $\eqref{eq:failure-classifier-loss}$ and $\eqref{eq:gp_margin}$. For the gradient penalty, we use parameters ${\lambda}_{\text{w}} = 0.1$, ${\lambda}_{\text{sign}} = 1$, ${\lambda}_{\text{GP}} = 10$ with the gradient threshold $\beta = 0.1$. These loss weights are chosen to prioritize $\mathcal{L}_{\text{sign}}$ and $\mathcal{L}_{\text{gp}}$. We choose a small weight for $\mathcal{L}_{\text{w}}$ since we observe that the gradient regularization can be lost when $\mathcal{L}_{\text{w}}$ is too high. For the baseline margin function without the gradient penalty, we only optimize $\mathcal{L}_{\text{sign}}^\delta$ with $\delta = 0.75$. 

We begin by training the diffusion policy: we collected 200 demonstrations using an MPPI controller with stochastic collision avoidance costs reaching a randomized goal. The initial conditions are concentrated between $x \in [-1.5,-1], y \in [-1, 1], \theta \in [-\frac{\pi}{3}, \frac{\pi}{3}]$ and goal location is fixed at $x = 1.3$ and the y position is sampled from $[-0.6, 0.6]$ This provides a dataset of task-oriented trajectories that are both safe and unsafe. To collect our world-model training data, we collect 4000 trajectories of randomly executed actions by sampling initial conditions uniformly over the state space and actions uniformly from the action space. We also collect 3800 trajectories of rolling out intermediate DP checkpoints at OOD initial conditions following \cite{sun2025latentpolicybarrierlearning}.%

\begin{table}[ht]
    \centering
    \small{
    \renewcommand{\arraystretch}{1.1}{
    \resizebox{0.5\linewidth}{!}{
    \begin{tabular}{lc}
        \toprule
        \textbf{\textsc{Hyperparameter}} & \textbf{\textsc{Value}} \\
        \midrule
        \textsc{RGB Image Dimension} & [128, 128, 3] \\
        \textsc{IR Image Dimension} & [128, 128, 1] \\
        \textsc{Action Dimension} & 1 \\
        \textsc{Stochastic Latent} & Gaussian \\
        \textsc{Latent Dim (Deterministic)} & 512 \\
        \textsc{Latent Dim (Stochastic)} & 32 \\
        \textsc{Activation Function} & SiLU \\
        \textsc{Encoder CNN Depth} & 32 \\
        \textsc{Encoder MLP Layers} & 5 \\
        \textsc{Failure Projector Layers} & 2 \\
        \textsc{Batch Size} & 32 \\
        \textsc{Batch Length} & 16 \\
        \textsc{Optimizer} & Adam \\
        \textsc{Learning Rate} & 1e-4 \\
        \textsc{Iterations} & 40000 \\
        \bottomrule
        \end{tabular}
        }
    }}
    \caption{Dubins' car RSSM Hyperparameters}
    \label{tab:dubins_hyperparams}
\end{table}

\begin{table}[h]
    \centering
    \begin{tabular}{l l}
        \toprule
        \textbf{Hyperparameter} & \textbf{Values}  \\
        \midrule
        State Normalization & Yes \\
        Action Normalization & Yes  \\
        Action Chunk & 16 \\
        Image Chunk & 2 \\
        Image Size & 256 \\
        Batch size & 100 \\
        Training Iterations & 500000 \\ 
        Learning Rate & 1e-4 \\
        Learning Rate Schedule & Cosine \\
        Optimizer & AdamW \\
    \bottomrule
    \end{tabular}
    \caption{Hyperparameters for Diffusion Policy}
    \label{tab:diffusion}
\end{table}

\para{Hardware: World Model and Nominal Policy} We train a nominal visuomotor diffusion policy $\piNom(\action \mid \obs)$ \citep{chi2024diffusionpolicy} using 215 teleoperated demonstrations and same hyperparameters as Table \ref{tab:diffusion}. In order to test the effectiveness of our safety filter in guiding an erroneous visuomotor policy, we intentionally include demonstrations of both \textit{unsafe} grasps from the closed-end of the bag (where the robot may spill) and \textit{safe} grasps from the opened end of the bag that prevents spilling. 
For our hardware experiments, we use DINO-WM \cite{zhou_dino-wm_2025}, a vision-transformer-based world model that predict future DINO embeddings conditioned on proprioception, robot action, and a history of $H=3$ timesteps. For this world model, we use frozen pre-trained \texttt{DINOv3-vits16plus} backbone \citep{simeoni2025dinov3} as our vision encoder. This compresses a $[256, 256, 3]$ RGB image into a $[256,384]$ tensor of dense patch token, capturing fine-grained detail in both the wrist and camera views. We train on  $|\mathcal{D}| = 735$ trajectories consisting of nominal policy rollouts from $\piNom$, random actions, and exploratory teleoperated trajectories following \citep{sun2025latentpolicybarrierlearning}.

We concatenate the patch tokens for both camera views, 10-d action embedding, and 8-dim state (eef position, quaternion, gripper width) along the feature dimension. This leads to a $[256, 786]$ latent representation of the state at each timestep. When training the world model, we normalize the dataset of actions to $\mathcal{N}(0,1)$ along each non-gripper action dimension. For ease of computation, we pre-compute the DINOv3 embeddings offline without image augmentations. This allows us bypass DINOv3 forward passes at each training iteration. Training the world model for 100000 iterations took about 12 hours on an Nvidia A6000 ADA GPU. We include all hyperparameters in Table \ref{tab:dinowm}.

To train the margin function and value function, we average-pool the latent state (average over patch tokens) before passing it as input to their respective MLPs. We parameterize $\ell(\latent)$ as a two-layer MLP with LayerNorm and ReLU activations. For the gradient penalty, we use parameters ${\lambda}_{\text{w}} = 0.2$, ${\lambda}_{\text{sign}} = 100$, ${\lambda}_{\text{GP}} = 10$ with the gradient threshold $\beta = 0.02$. These loss weights are chosen to prioritize $\mathcal{L}_{\text{sign}}$ and $\mathcal{L}_{\text{gp}}$. For the baseline margin function without the gradient penalty, we once again we only optimize $\mathcal{L}_{\text{sign}}^\delta$ with $\delta = 0.75$.  

\begin{table}[h]
    \centering
    \begin{tabular}{l l}
        \toprule
        \textbf{Hyperparameter} & \textbf{Values}  \\
        \midrule
        Image size & 256 \\
        DINOv3 patch size & $(16 \times 16, 384)$ \\ 
        Optimizer & AdamW \\
        Predictor lr & 5e-5  \\
        Decoder lr & 3e-4 \\
        Action Encoder lr & 5e-4 \\
        Action emb dim & 10 \\
        Proprioception emb dim & 10 \\
        Batch size & 16 \\
        Batch len & 4 \\
        Training iterations & 100000 \\
        ViT depth & 6 \\
        ViT attention heads & 16 \\
        ViT MLP dim & 2048 \\
    \bottomrule
    \end{tabular}
    \caption{Hyperparameters for DINO-WM}
    \label{tab:dinowm}
\end{table}

\subsection{Latent Space Reinforcement Learning}
\label{subsec:latent-RL}
We list all parameters for our actor and critic networks in Table \ref{tab:DDPG_hyperparams}. We parameterize our safety policy $\fallback$ as a deterministic policy and use a single critic. During RL training, we assume access to a policy $\piNom$ that takes as input image observations. In practice, we use a diffusion policy that outputs an H-timestep action chunk $a_{t:t+H}$ for H=16 steps. 

To perform reinforcement learning, one has to "reset" the environment at the beginning of each trajectory. Due to the high-dimensionality of the latent state, this reset must be done carefully to ensure resetting onto the data manifold. In practice, this is done by encoding a randomly sampled observation from an offline dataset $\obs \sim \dataset$. We generate 
$a_{t:t+H} \sim \piNom(\obs)$ with 50$\%$ probability and execute the first $T$ timesteps before terminating the episode. We do not explicitly require this form of nominal policy, in principle one could repeat a single action $\action \sim \piNom(\obs)$ for $T$ steps for other parameterizations of the nominal policy. Otherwise, we execute $\fallback$ the entire $T$ timestep episode. In practice, we set $T=8$ to prevent learning on world model imaginations that degrade for long time horizons. The actor is constrained to output actions within 
$[-1,1]$ per dimension, which on hardware corresponds approximately to $\pm 1$ standard deviation of the dataset actions, given that the world model operates under normalized action scaling. When storing the $(z, a, l, z, a')$ tuple into the dataset we use $l = \tanh(\ell(\latent))$ to ensure boundedness.

\begin{table}[h]
    \centering
    \resizebox{0.5\linewidth}{!}{
    \renewcommand{\arraystretch}{1.1}
    \begin{tabular}{lc}
        \toprule
        \textsc{\textbf{Hyperparameter}} & \textsc{\textbf{Value}} \\
        \midrule
        \textsc{Actor Architecture} & [512,  512, 512] \\
        \textsc{Critic Architecture} & [512, 512, 512] \\
        \textsc{Normalization} & LayerNorm \\
        \textsc{Activation} & ReLU \\
        \textsc{Discount Factor} $\gamma$ & 0.995 \\
        \textsc{Learning Rate (Critic)} & 3e-4 \\
        \textsc{Learning Rate (Actor)} & 1e-4 \\
        \textsc{Optimizer} & Adam \\
        \textsc{Number of Iterations} & 120000 \\
        \textsc{Replay Buffer Size} & 100000  \\
        \textsc{Batch Size} & 512 \\
        \textsc{Max Imagination Steps} & 8 \\
        \bottomrule
    \end{tabular}
    }
    \caption{RL hyperparameters.}
    \label{tab:DDPG_hyperparams}
    \vspace{-0.in}
\end{table}

\subsection{Proof of Margin-to-Value Lipschitz Bound}
\label{subsec:app-proof-bound}
For clarity, we restate the statement we wish to prove.
\begin{theorem}[Margin-to-Value Lipschitz Bound]
    Let the margin function $\marginfunc(\state)$ and time discounted value function $\valfunc(\state)$ be Lipschitz continuous with constants $L_{\marginfunc}$ and $L_{\valfunc}$,  respectively. Let the discrete-time dynamics $\dyns(\state, \action)$ be uniformly Lipschitz in $\state$ with constant $L_\dyns$ such that for a fixed discount factor $\gamma \in [0,1)$, $\gamma L_\dyns < 1$.
    Then the Lipschitz constant of $V(\state)$ scales linearly in $L_\marginfunc$ :%
    \[ 
    L_{\valfunc} \leq L_\marginfunc \cdot \max \left\{1, \frac{1-\gamma}{1 - \gamma L_\dyns} \right\}. 
    \]
\end{theorem}

\noindent\textbf{Proof:} We seek an upper bound on $L_{\valfunc}$ which is defined as the smallest constant such that $\forall \state, \statet \in \stateSpace$,
$|\valfunc(\state) - \valfunc(\statet)| \leq L_{\valfunc} |\state - \statet|$. We begin by expanding $\valfunc(\state) - \valfunc(\statet)$.

\begin{equation*}
\small
\begin{aligned}
        \valfunc(\state) - \valfunc(\statet) &= \\ &(1-\gamma)(\marginfunc(\state) -\marginfunc(\statet)) + \gamma \big[ \min \Big\{ \marginfunc(\state), ~ \max_{\action \in \actionSpace} \valfunc(\dyns(\state, \action)) \Big\} - \min \Big\{ \marginfunc(\statet), ~ \max_{\actiont \in \actionSpace} \valfunc(\dyns(\statet, \actiont)) \Big\} \big] 
\end{aligned}
\end{equation*}

\para{Case 1} $\marginfunc(\statet)$ minimizes the second term.
\begin{equation*}
\begin{aligned}
        \valfunc(\state) - \valfunc(\statet) \leq (1-\gamma)(\marginfunc(\state) -\marginfunc(\statet)) + \gamma(\marginfunc(\state) - \marginfunc(\statet)) \end{aligned}
\end{equation*}
since $\min \Big\{ \marginfunc(\state), ~ \max_{\action \in \actionSpace} \valfunc(\dyns(\state, \action)) \Big\} - \marginfunc(\statet) \leq \marginfunc(\state) - \marginfunc(\statet)$.

\para{Case 2} $\max_{\actiont \in \actionSpace} \valfunc(\dyns(\statet, \actiont)) $ minimizes the second term.
\begin{equation*}
\begin{aligned}
        \valfunc(\state) - \valfunc(\statet) \leq (1-\gamma)(\marginfunc(\state) -\marginfunc(\statet)) + \gamma(\valfunc(\dyns(\state, \action')) - \valfunc(\dyns(\statet, \action')))
\end{aligned}
\end{equation*}
where $a' = \text{argmax}_{\action \in \actionSpace} \valfunc(\dyns(\state, \action))$ and we use similar logic to Case 1 to upperbound the difference.

Note that $\valfunc(\statet) - \valfunc(\state)$ yields symmetric bounds. This implies that we are left with two cases:

\begin{equation*}
\begin{aligned}
        |\valfunc(\state) - \valfunc(\statet)| \leq |\marginfunc(\state) -  \marginfunc(\statet)| \leq L_\marginfunc|s - s'|
\end{aligned}
\end{equation*}

and 

\begin{equation*}
\begin{aligned}
        |\valfunc(\state) - \valfunc(\statet)| \leq & (1-\gamma)L_\marginfunc|s - \tilde{s}| + \gamma L_{\valfunc}|\dyns(\state, \action') - \dyns(\statet, \action')| \\ \leq & (1-\gamma)L_\marginfunc|s - \tilde{s}| + \gamma L_{\valfunc} L_\dyns |\state - \statet| = ((1-\gamma)L_\marginfunc + \gamma L_{\valfunc} L_\dyns)|\state - \statet|
\end{aligned}
\end{equation*}
Since by definition $L_{\valfunc}$ is the smallest constant $L$ such that $|\valfunc(\state) - \valfunc(\statet)| \leq L |\state - \statet |$, rearranging results in
\begin{equation*}
\begin{aligned}
        L_{\valfunc} \leq \max\{L_\marginfunc, \frac{1-\gamma}{1 - \gamma L_\dyns } L_\marginfunc \}.
    \end{aligned}
\end{equation*}
In both cases, $L_{\valfunc}$ varies linearly in $L_\marginfunc$. While the assumptions made have no guarantee to hold in latent spaces induced by a world model (whose continuity properties are poorly understood in general), or may lead to weak bounds if the Lipschitz constant of the dynamics is high (e.g., due to contacts), this motivates restricting the Lipschitz constant\footnote{The only assumption we required of $\marginfunc(\latent)$ is boundedness for the existence and uniqueness of the value function. However, under simple neural network parameterizations this function is indeed Lipschitz continuous in practice.} of $\marginfunc(\latent)$ to improve the gradient landscape of the value function.

\subsection{Additional Simulation results}

We report the confusion matrix (where TP corresponds to classified as safe and ground truth is safe, etc.) for margin function accuracy, and our smoothness metric in Table \ref{tab:lz}. The gradient penalty drastically reduces the largest $\Delta_t \ell(\latent_t)$ without significantly degrading classification accuracy.

\begin{table}[ht!]
\centering
\setlength{\tabcolsep}{3pt}
\renewcommand{\arraystretch}{0.8}
\begin{tabular}{l|ccccc}
\toprule
$\marginfunc{_\ellparam}(\latent)$ & TP (\%) & TN (\%) & FP (\%) & FN (\%) &  $\max_t \Delta_t\marginfunc(\latent_{t})$  \\
\midrule
NoGP &  84 & 13 & 0.63 & 2.5 & 1.2 $\pm$ 0.76  \\
GP & 86 & 13 & 1.0 & 0.60 & 0.17 $\pm$ 0.065 \\
\bottomrule
\end{tabular}
\caption{\textbf{Simulation: Accuracy \& Smoothness of Latent $\marginfunc{_\ellparam}(\latent)$.} In the Dubins' car simulation, we compare the latent failure set encoded via $\failure_\latent = \{\latent : \marginfunc{_\ellparam}(\latent) < 0\}$ to the ground-truth failure set, $\failure \subset \stateSpace$, as well as the maximum $\marginfunc{_\ellparam}(\latent)$ gradient along trajectories generated by $\piNom$ with no filtering, averaged across 100 initial conditions. }
\label{tab:lz}
\vspace{-1.8em}
\end{table}

\para{Ablating $\alpha$}
We ablate the choice of $\alpha$ on our simulated Dubins' example in Section \ref{sec:sim}. We note that for the NoGP baseline, the CBF is not sensitive to lower values of $\alpha$ while the gradient penalty allows us to tune the aggressiveness of the filter (as shown by $|\Delta \action|$).

{
\renewcommand{\arraystretch}{0.5}
\begin{table}[h!]
\centering
\setlength{\tabcolsep}{3pt}
\begin{tabular}{l|cc|cc|c}
\toprule
\multirow{3}{*}{\textbf{Safety Filter}} 
& \multicolumn{2}{c|}{\textbf{NoGP}} 
& \multicolumn{2}{c|}{\textbf{GP}} 
& \textbf{None} \\
\cmidrule(lr){2-3} \cmidrule(lr){4-5} \cmidrule(lr){6-6}
& Avg. ${|\Delta\action|}$ & Safety Rate 
& Avg. ${|\Delta\action|}$ & Safety Rate 
& Safety Rate \\
\midrule
LR   & $1.5\pm1.1$ & $97\%$ & $2.0\pm1.3$ & $100\%$ & \multirow{3}{*}{$41\%$} \\
CBF ($\alpha=0.7$)  & $1.3\pm1.1$ & $96\%$ & $1.3\pm1.1$ & $100\%$ &  \\
CBF ($\alpha=0.95$) & $1.3\pm1.0$ & $97\%$ & $1.1\pm0.9$ & $100\%$ &  \\
\bottomrule
\end{tabular}
\caption{\textbf{Simulation: On the Quality of Latent Safety Filters.}}
\label{tab:cbf_dubins}
\vspace{-0.5cm}
\end{table}
}

\begin{figure}[hbt!]
    \centering
    \includegraphics[width=0.9\linewidth]{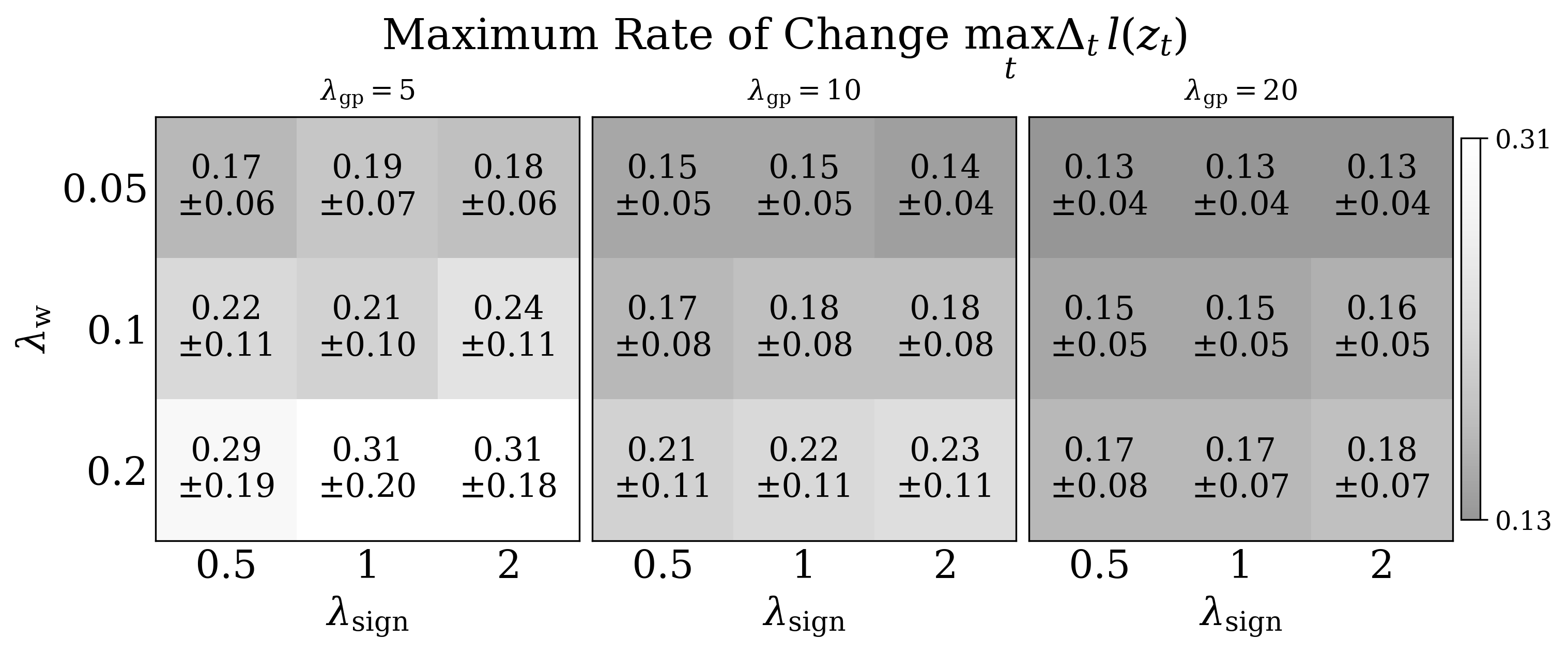}
    \caption{Sensitivity Study: Smoothness}
    \label{fig:ablation}
\end{figure}

\begin{figure}[hbt!]
    \centering
    \includegraphics[width=0.9\linewidth]{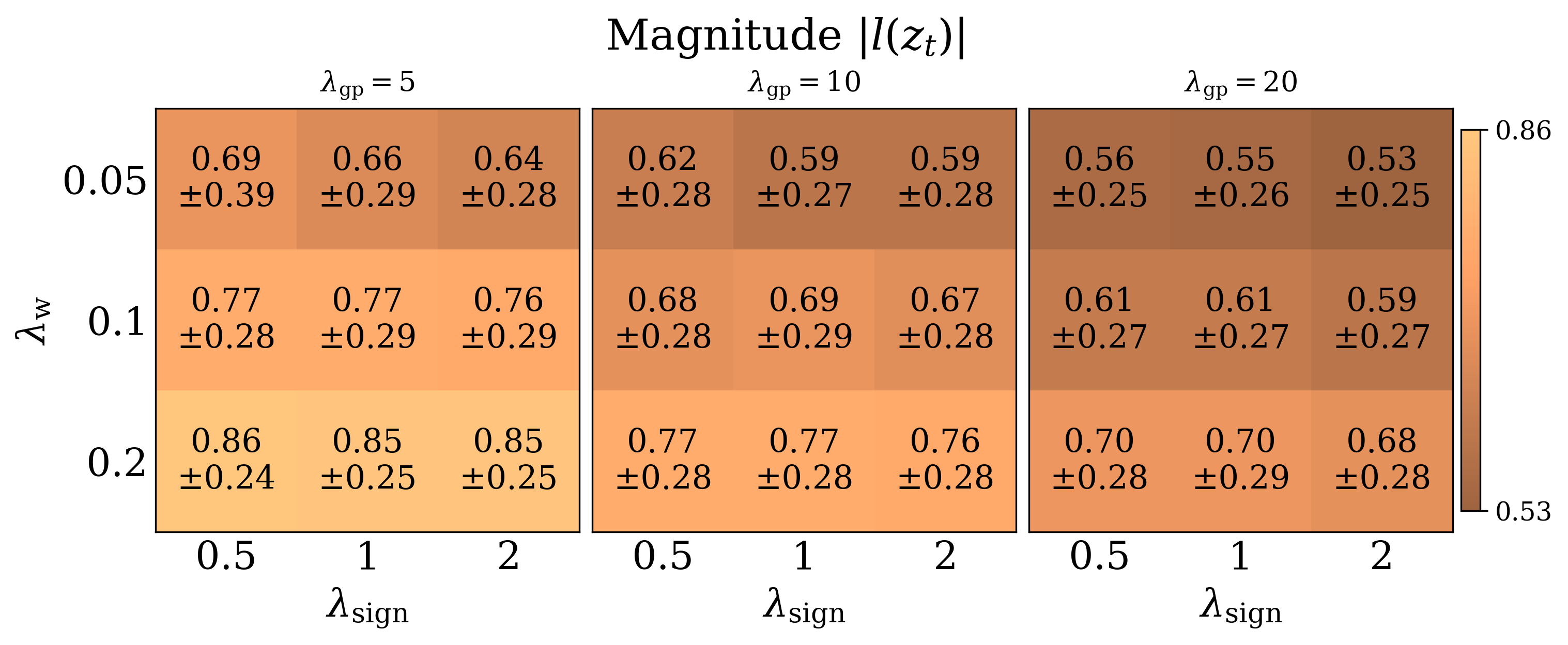}
    \caption{Sensitivity Study: Margin magnitude.}
    \label{fig:mag}
\end{figure}

\begin{figure}[hbt!]
    \centering
    \includegraphics[width=0.9\linewidth]{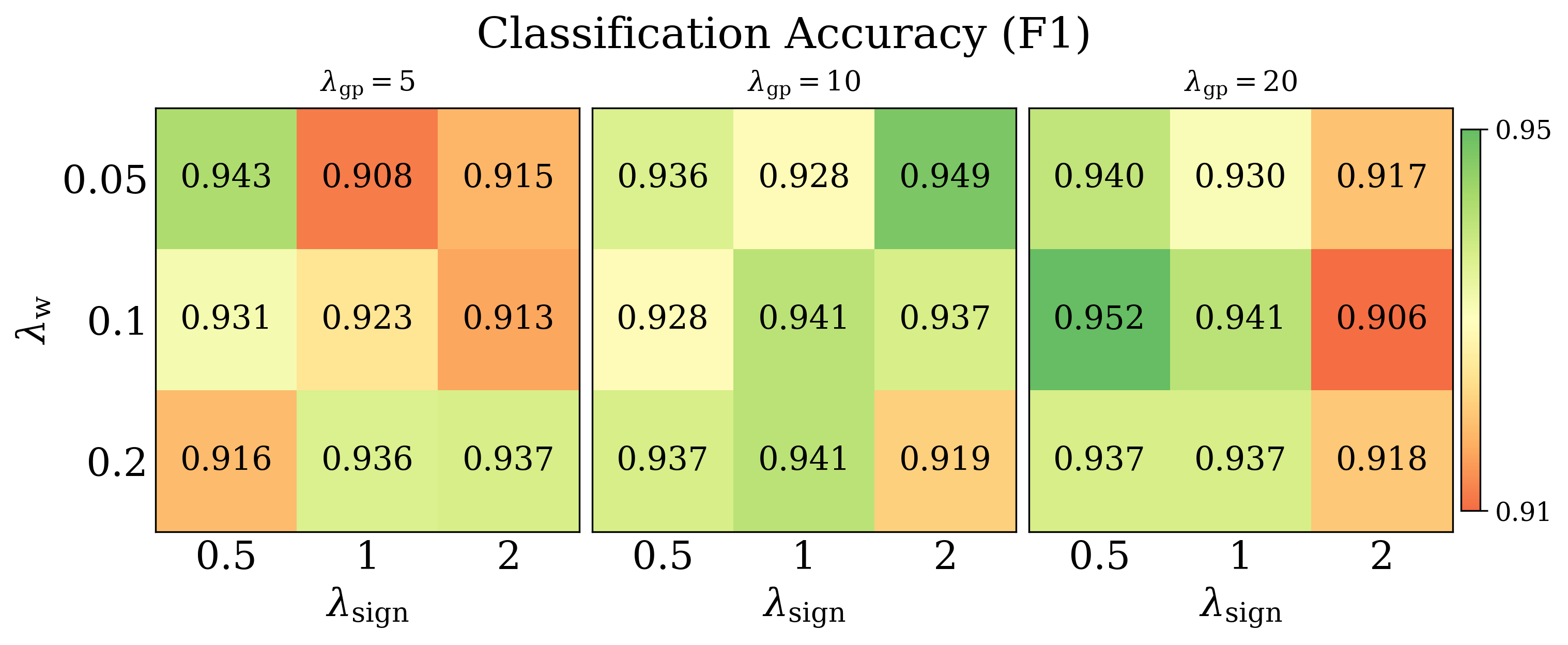}
    \caption{Sensitivity Study: Classification Accuracy.}
    \label{fig:f1}
\end{figure}

\para{Hyperparameter Selection}
Our WGAN inspired method for learning a smooth margin function has several weighting terms. At a high level, $\lambda_{\text{w}}$ dictates encourages safe states to take large positive values and unsafe states to take large negative values. However, since this does not semantically encode that $\marginfunc(\latent) < 0 \iff \latent \in \failure$, we use  $\lambda_{\text{w}}$ to weigh how much we require the margin function to abide by this constraint. To discourage the Lipschitz constant of this network from growing too large, we add a gradient penalty \citep{gulrajani2017improved} with weight $\lambda_{\text{gp}}$ and Lipschitz constant target $\beta$. 
We analyze the margin function's sensitivity with respect to a range of $\lambda_{\text{w}}$, $\lambda_{\text{sign}}$, and $\lambda_{gp}$ in Figures \ref{fig:ablation}, \ref{fig:mag}, and \ref{fig:f1} using 100 trajectories collected by the base task policy. 


We consistently find that increasing $\lambda_{\text{w}}$ increases average magnitude $|l(z_t)|$ and the maximum single-timestep change $\max_t \Delta_t l(z_t)$, while these metrics are relatively insensitive to choice of $\lambda_{\text{sign}}$. Increasing $\lambda_{\text{gp}}$ generally reduces the sensitivity of $\max_t \Delta_t l(z_t)$ with respect to $\lambda_{\text{w}}$ and lowers the magnitude $|l(z_t)|$. On the other hand, the F1 score is consistently high across a range of hyperparameters, supporting our claim that the WGAN-GP training objective \eqref{eq:wgan} is independent of additive shifts of $\marginfunc(\latent)$. In practice, we tuned the hyperparameters in the main body by fixing $\lambda_{\text{gp}} = 10$ and selecting the target Lipschitz constant $\beta$ according to the average gradient norm at the beginning of training. We then tuned $\lambda_{\text{w}}$ so that the maximum magnitude of the margin function did not greatly exceed $1$, (as we bound the margin function between $[-1,1]$, as noted in Section \ref{subsec:latent-RL}). Afterwards, $\lambda_{\text{sign}}$ was tuned so the resulting margin function had high classification accuracy.

\subsection{Sampling-Based Optimization of the Latent Discrete-time CBF}
\label{sec:sampling_appendix}
Unlike the continuous-time CBF with control-affine dynamics, the discrete-time CBF optimization problem in \eqref{eq:latent_cbf} does \textit{not} admit a quadratic program and is in general nonconvex, making it challenging to optimize efficiently in high-dimensional action spaces. We use a zeroth-order optimization procedure, accelerated with the parallelization capabilities of modern hardware.
Specifically, we sample a large number of actions in parallel (e.g., in hardware its 7,600) from a mixture of the nominal task-driven policy $\piNom$ and the safety policy $\fallback$ to form a set of actions $\actionSpace_{\text{sample}} \subset \actionSpace$. 
We then find a subset of the samples which satisfy the CBF constraint 
\begin{equation}
\actionSpace_{\text{CBF-Safe}}=\{\action_i \in \actionSpace_{\text{sampled}} ~|~(\qfunc(\latent,\action_i) - \epsilon) \geq \alpha (\valfunc(\latent) - \epsilon)\}    
\end{equation} 
by evaluating the learned $\qfunc(\state,\action_i)$ for each sampled action $\action_i$ and comparing it to the safety value assuming we switched to the safety-preserving policy, $\valfunc(\latent) = \qfunc(\latent, \fallback(\latent))$. Here the hyperparameter $\epsilon$ is a small positive constant which accounts for learning inaccuracies of the zero-level set. 
Finally, we execute actions via: \begin{equation}
\label{eq:latent_filter}
        \action^\star =
        \begin{cases} 
        \fallback(\enc(\obs)) & \textbf{if } \actionSpace_{\text{CBF-Safe}} = \emptyset \\
          \argmin_{\action \in \actionSpace_{\text{CBF-Safe}}}||\action-\piNom(\obs)|| & \textbf{else} \\
   \end{cases}
    \end{equation}
where $\fallback$ is executed in the case where none of the action samples satisfy the discrete-time CBF constraint; otherwise, we return the most similar action sample from $\actionSpace_{\text{CBF-Safe}}$ to the nominal policy by evaluating all samples in parallel (e.g., in hardware, the overall process of finding $\actionSpace_{\text{CBF-Safe}}$ and evaluating the nearest action takes about $10$ ms for 7.6k samples for our 7DOF manipulator).

\subsection{Sampling Distributions} 
\label{subsec:sampling}
For our simulated scenarios, we sample $N=25$ equally spaced action samples $\action \in [-2, 2]$ at each timestep. We also include the actions from $\fallback$ and $\piNom$ for a total of 27 action samples. 

For our hardware manipulation example, we have a 7-dimensional action space (delta position, axis angle rotation, gripper open/close) which is intractable to sample over exhaustively. Instead, we chose our sampling scheme by carefully linearly interpolating between key vectors over specific action dimension. We fully describe the sampling process in Table \ref{tab:sample}. For each set of samples we interpolate the specified subset of action dimensions (for example just x action, or x, y, and z) between two actions while holding the remaining dimensions at the action specified by \textbf{Static}. For example, while the top row results in a $(400,7)$ action tensor that interpolates all actions between actions returned by $\piNom$ and $\fallback$, the second row only interpolates along the $(\Delta x, \Delta y, \Delta z)$ components while keeping the rotation and gripper actions fixed to that of $\piNom$.

\begin{table}[h!]
\centering
\begin{tabular}{c|c|c|c|c}
\toprule
\textbf{\#} & \textbf{Interp. Dims} & \textbf{Interp. From} & \textbf{Interp. To} & \textbf{Static} \\
\midrule
1 & $[\mathbf{x},\mathbf{y},\mathbf{z},\mathbf{\omega_x},\mathbf{\omega_y},\mathbf{\omega_z}]$ & \multirow{10}{*}{{$\piNom$}} & \multirow{6}{*}{{$\fallback$}} & \multirow{10}{*}{{$\piNom$}}\\
2 & $[\mathbf{x},\mathbf{y},\mathbf{z}]$ & & & \\
3 & $[\mathbf{\omega_x},\mathbf{\omega_y},\mathbf{\omega_z}]$ & & \\
4 & $[\mathbf{x}]$ & & \\
5 & $[\mathbf{y}]$ & & \\
6 & $[\mathbf{z}]$ & & \\
\cmidrule(lr){2-2} \cmidrule(lr){4-4}
7 & $[\mathbf{x},\mathbf{y},\mathbf{z},\mathbf{\omega_x},\mathbf{\omega_y},\mathbf{\omega_z}]$ & & \multirow{4}{*}{\textbf{$\mathbf{0}$}} \\
8 & $[\mathbf{x}]$ & & \\
9 & $[\mathbf{y}]$ & & \\
10 & $[\mathbf{z}]$ & & \\
\cmidrule(lr){2-5}
11 & $[\mathbf{x}]$ & \multirow{6}{*}{$\fallback$} & \multirow{3}{*}{{$\piNom$}} & \multirow{6}{*}{$\fallback$} \\
12 & $[\mathbf{y}]$ & & \\
13 & $[\mathbf{z}]$ & & \\
\cmidrule(lr){2-2} \cmidrule(lr){4-4} 
14 & $[\mathbf{x}]$ & & \multirow{3}{*}{$\mathbf{0}$} \\
15 & $[\mathbf{y}]$ & & \\
16 & $[\mathbf{z}]$ & & \\
\cmidrule(lr){2-5}
17 & $[\mathbf{x}]$ & \multirow{3}{*}{$\bm{\upmu-\upsigma}$} & \multirow{3}{*}{$\bm{\upmu+\upsigma}$} & \multirow{3}{*}{$\piNom$} \\
18 & $[\mathbf{y}]$ & & \\
19 & $[\mathbf{z}]$ & & \\
\bottomrule
\end{tabular}
\caption{Scheme for generating action samples for the manipulator. Each linear interpolation takes $N=400$ samples for a total of $7600$ samples. Each line is interpolated by first deciding which set of actions to interpolate from and to, as well as which dimensions participate in the interpolation; while the rest are held static. Bounds $\bm{\upmu+\upsigma}$ and $\bm{\upmu-\upsigma}$ represent actions $+1$ and $-1$ standard deviation away from the mean actions from the dataset.}
\label{tab:sample}
\end{table}

\end{document}